# Exploring Academic Influence of Algorithms by Co-occurrence Network Based on Full-text of Academic Papers

Yuzhuo Wang[1], Chengzhi Zhang[2, *], Min Song[3], Seong Deok Kim[3], Youngsoo Ko[3], Juhee Lee[3]

1. Department of Management Science and Engineering, Anhui University, Hefei, 230601, China

2. Department of Information Management, Nanjing University of Science and Technology, Nanjing, 210094, China

3. Department of Information Science, Yonsei University, Seoul, 03722, Korea

**Abstract**

**Purpose-** In the era of artificial intelligence (AI), algorithms have gained unprecedented importance. Scientific studies have shown that algorithms are frequently mentioned in papers, making mention frequency a classical indicator of their popularity and influence. However, contemporary methods for evaluating influence tend to focus solely on individual algorithms, disregarding the collective impact resulting from the interconnectedness of these algorithms, which can provide a new way to reveal their roles and importance within algorithm clusters. This article aims to build the co-occurrence network of algorithms in the natural language processing (NLP） field based on the full-text content of academic papers, and analyze the academic influence of algorithms in the group based on the features of the network.

**Design/methodology/approach-** We use deep learning models to extract algorithm entities from articles and construct the whole, cumulative, and annual co-occurrence networks. We first analyze the characteristics of algorithm networks, and then use various centrality metrics to obtain the score and ranking of group influence for each algorithm in the whole domain and each year. Finally, we analyze the influence evolution of different representative algorithms.

**Findings-** The results indicate that algorithm networks also have the characteristics of complex networks, with tight connections between nodes developing over approximately two decades. For different algorithms, Algorithms that are classic, high-performing, and appear at the junctions of different eras can possess high popularity, control, central position, and balanced influence in the network. As an algorithm gradually diminishes its sway within the group, it typically loses its core position first, followed by a dwindling association with other algorithms.

**Originality/value-** This article is the first large-scale analysis of algorithm networks. The extensive temporal coverage, spanning over four decades of academic publications, ensures the depth and integrity of the network. Our results serve as a cornerstone for constructing multifaceted networks interlinking algorithms, scholars, and tasks, facilitating future exploration of their scientific roles and semantic relations.

**Keywords:** Algorithm co-occurrence network, Algorithm entities, Full-text content of Articles

## 1. Introduction

We currently stand at the dawn of the era of artificial intelligence (AI). The AI revolution is reshaping our lives at an astonishing pace, fundamentally altering how we work, learn, and even live together (Azoulay, 2019). In this new era, algorithms, computing power, data, and knowledge are the key elements of artificial intelligence (Tang, 2021; Byte Bridge, 2021; Deng *et al*., 2020), with algorithms being particularly significant because they are the backbone of computer systems and are essential in revealing "indirect, non-obvious" corrections in data (Hickman, 2013). In today's data-driven research landscape, algorithms play a critical role, as they are important research objects and have been optimized, innovated, and proposed by scientists to cater to the evolving demands of data-driven exploration (Balcan, 2020; Gupta and Roughgarden, 2020). At the same time, in data-driven research, information technology is engaged in the operation, collection, transport, retrieving, storage, access presentation, and transformation of information in all its forms, inseparable from various algorithms. Indeed, algorithms are the driving to solve real-world challenges, from monitoring epidemics (Kogan *et al.*, 2021) to enhancing intelligent transportation (Matheson, 2020), optimizing agricultural management (Talaviya *et al.*, 2020), and enabling smart homes (Fleury *et al.*, 2010).

[*] Corresponding author, Email: zhangcz@njust.edu.cn

However, defining algorithms remains a nuanced and often contentious endeavor (Blass and Gurevich, 2004), and usually depends on whom we ask (Lum and Chowdhury, 2021). In the general context, an algorithm is a set of rules that precisely define a sequence of operations (Stone, 1971). In the context of programming, some scholars equated algorithms to a programming language (Knuth, 1996). Cormen *et al*. (2009) provided a more detailed definition: "Informally, an algorithm is any well-deified computational procedure that takes some value, or set of values, as input and produces some value, or set of values, as output." Within the AI realm, an algorithm was recognized as "a method of analyzing data in a way that makes it possible for the machine to analyze and process data" (Genics, 2019).

While the definitions of algorithms may vary, it is irrefutable that algorithms are frequently mentioned in academic papers, making academic papers a crucial medium for disseminating algorithm-related knowledge (Bhatia and Mitra, 2012). Scholars directed their attention towards extracting algorithm-related metadata (Tuarob, 2016) and algorithm entities (Wang and Zhang, 2020) from academic papers, and evaluating their academic influence or performance to form a leaderboard ("Papers with Code - Browse the State-of-the-Art in Machine Learning", 2023; "The Heart of Machine", 2023). The academic influence of algorithms serves as a pivotal reference point for scholars when selecting and understanding algorithms, and sheds light on the development trends within a specific field (Wang and Zhang, 2020). However, the current academic influence evaluation only focuses on individual algorithm characteristics, neglecting the relationship between algorithms and the role of algorithms in the community, which could provide a new perspective on measuring individual influence from the group function. Consider two algorithms with an identical frequency of mentions, and they may have different statuses within a community. Regrettably, there has been no research analyzing whether the algorithms in a field have been mentioned together or exploring the interconnection, separation, status gap, and other attributes between the algorithm individuals in a community. To fill this gap, we construct a co-occurrence network of algorithms and measure the influence based on the centrality indicators, shedding light on the broader context of how algorithms are employed in scientific research.

Taking the algorithms in natural language processing (NLP) as research objects, we sought to answer the following research questions:

***RQ1.*** *What is the structure of the NLP algorithm network, and how has it evolved over time?*

We want to understand if the network constructed by algorithms also has the same properties as other complex networks, how the current algorithm network was formed, and whether the connection between algorithms changed during the development of the NLP field.

***RQ2***. *What are the vital algorithms in the algorithm network?*

Various centralities in complex networks have been used to measure the influence of countries, institutions, scholars, papers, keywords, terms, and other scientific objects. However, there has been little research on the role and influence of algorithm entities based on community features. After confirming that the algorithm network conforms to the complex network characteristics, this paper aims to use different indicators within the network, identify representative algorithms in various roles, and provide new insights into evaluating academic influence.

***RQ3***. *How does the academic influence of algorithms based on the network features change?*

We aim to discern whether algorithms experience shifts in influence and roles within the network, and manifest distinctive attributes at different stages of its life cycle.

The research object in this article is the algorithm entity, not the metadata (the pseudo-code), types of algorithms (classification algorithms), or general algorithms (Bubble sorting). Based on existing research (Cormen *et al*., 2009; Genics, 2019; Wang *et al*., 2022; Zhang *et al*., 2023), we define an algorithm in our study as a well-deified computational procedure with a specific name, like support vector machines. It should be distinguished via software (VOSviewer), system (Unix), dataset (ACL RD-TEC), and metrics (BLEU), while an algorithm entity is a noun or noun phrase indicating the name of an algorithm. Without special instructions, both *algorithms* and *algorithm entities* refer to nouns or noun phrases representing explicit algorithms.

Unlike academic object networks that rely solely on bibliographic information, the networks in this study were constructed using the full-text content of scholarly articles, since the full text enables us to collect more algorithm nodes, obtain a complete network, and produce a landscape of the algorithms in the NLP domain. Generally, our work lays the foundation for constructing networks based on fine-grained knowledge entities and using network characteristics to analyze their influence, explore their relationships, and analyze their evolution. Based on algorithm entities, our research provides a comprehensive methodology that spans from entity extraction to network construction and then impact evaluation, which is different from traditional counting methods and allows for distinguishing the roles of algorithms within a group. It offers a reference for measuring the influence and status of other types of knowledge in different fields. The influence evaluation results supplement the traditional algorithm leaderboard, revealing key algorithms that hold significant positions and play a driving role in the algorithm community of the entire field. Therefore, our results can be applied by researchers to understand the history of algorithms, choose algorithms for research, and plan the development of methodologies in a field. Finally, based on the analysis of the algorithm co-occurrence network presented in this paper, future work could further reveal the relationships and similarities between algorithms, potentially aiding the development of algorithm recommendation and retrieval services.

## 2. Related work

This article focuses on a network comprised of the algorithms mentioned in academic papers. In this section, we introduce three aspects of related work: algorithms mentioned in academic papers, academic influence evaluation, and the various networks of knowledge entities.

## 2.1 Algorithms mentioned in academic papers

The algorithms mentioned in academic papers are currently identified to conduct further research. Relevant studies can be classified into algorithm metadata research and algorithm entity research, according to the different expressions of algorithms.

Algorithm metadata refers to algorithm-related pseudo-code and algorithm procedures. For algorithms represented by pseudo-code, textual metadata includes their captions and reference sentences; for algorithms described by algorithm procedures, their textual metadata includes indication sentences (Tuarob, 2014). Safder *et al*. conducted a series of investigations into algorithm metadata; specifically, they tried various methods of identifying algorithm metadata in academic papers, including a rule-based method (Bhatia *et al.*, 2011), a combined rule-based and machine learning method (Tuarob *et al*., 2013a), and a deep learning method (Safder *et al.*, 2020). The cited function of algorithm metadata was further analyzed (Tuarob *et al*., 2019; Tuarob *et al*., 2013b). Subsequently, they built a platform for algorithm searching, called AlgorithmSeer*,* which provides algorithm content for users by calculating the similarity between the algorithm metadata and the searched terms (Carman, 2013; Tuarob *et al.*, 2014, 2016).

An algorithm entity is a noun or noun phrase that refers to an algorithm in academic papers. Algorithm entity research focuses on algorithm entity recognition and influence evaluation. Wang and Zhang (2018) built a dictionary of the top ten data mining algorithms, obtained the context mentioning the ten algorithms, and then used the frequency, location, and task solved by the algorithm to evaluate the influence of data mining algorithms. On this basis, they summarized the sentence pattern for mentioning the data mining algorithms, and used this approach to identify other types of method entities from sentences (Wang and Zhang, 2019). Given that the top-ten data mining algorithms cannot reveal the algorithm landscape in a particular domain, Wang and Zhang (2020) extended the research object to the algorithm entities in the NLP domain from 1979 to 2015, and evaluated the academic influence of the algorithm via the number of articles and the duration of the mention. Ding *et al*. (2019) further analyzed the citation sentences of algorithm entities, and discussed the evolution of algorithm citation sentences over time, where algorithm entities were analyzed as individual objects. Zha *et al*. (2019) further considered the relationships between algorithms, and extracted a comparative algorithm list from academic papers. Next, they built the roadmap of the compared algorithms according to the time they were mentioned.

### 2.2 Academic influence evaluation on knowledge entities

Distinct from social influence, academic influence emphasizes the insights and cognitive impacts brought to researchers when the evaluated subject circulates within the academic, rather than the social domain. Academic influence engages the audience, prompting their attention and contemplation, altering their intellectual thinking and actions, and inspiring academic innovation.

The evaluation of academic influence on knowledge entities traditionally relies on qualitative assessments, often conducted through peer review. Scholars typically establish an assessment system or framework and then collect evaluation results for specific entities using expert assessments, interviews, questionnaires, etc. A notable representative work is the selection of the top 10 data mining algorithms (Wu et al., 2018). Similarly, the Stack Overflow 2023 Global Developer Report invited 90,000 IT professionals to participate in a survey and identify the world's most popular technologies, web frameworks, and databases (Stack Overflow, 2023). The TIOBE index, calculated based on the number of votes from experienced internet programmers, courses, and third-party vendors, ranks the popularity of different programming languages (TIOBE-index, 2023). These rankings offer references for scholars, but they require a considerable workforce and are unsuitable for frequent, long-term implementation.

Scholars proposed various quantitative indicators to compensate for the shortcomings of qualitative evaluation methods. Citation count is the most commonly used metric for measuring academic influence (He et al., 2019). Belter and I (2014) focused on three datasets in the field of oceanography, studying the impact of datasets based on citation count. With the availability of full-text data, the number of mentions and usage of knowledge entities in full texts are used to assess their influence. Pan et al. (2015) used a bootstrapped learning method to extract software entities automatically and assessed their influence based on citation count, usage count, and mention count. They further compared the academic impact of different categories of software entities (Pan et al., 2016) and analyzed the influence diffusion trends of software entities at the levels of literature, journals, and fields (Pan et al., 2018). Furthermore, Ma and Zhang (2017) used total mentions and average mentions to assess the academic influence of software entities, identifying typical roles of software in PLOS One. In addition to various mention metrics, altermetrics indicators have also been used to measure the influence of knowledge entities (Li and Xu, 2017). Thelwall and Kousha (2016) considered download counts of open-source software to be the assessment metric of software value, which Priem and Piwowar then used to initiate the Depsy project ( "Depsy:valuing the software that powers science", 2023), aiming to analyze the influence of software in the open-source community by download counts and citation counts. Some scholars attempted to combine other attributes with frequency for influence analysis. Tuarob et al. explored the functions mentioned of algorithmic entities in papers (Teufel, 2006; Tuarob et al., 2019; Tuarob, Mitra, et al., 2013), while Yoon *et al.*, (2019) analyzed the sources of dataset influence. Li et al. (2020) analyzed position and citation circumstances of datasets in full-text papers for a comprehensive assessment of their academic influence.

**2.3 Evolution of knowledge entity networks**

In 2003, Ding *et al*. (2013) proposed entity metrics in which the individual knowledge units encapsulated as entities in scientific papers can be connected to mine knowledge transfer and facilitate knowledge discovery. Network analysis is a standard method for processing connected data, which is also used to construct knowledge entity networks and study their structure and evolution mechanism. According to the objects analyzed, knowledge entity networks can be divided into three levels: macro-level entity networks, including journal networks (Barnett *et al.*, 2011), author networks (Kumar, 2015), and reference networks (Chen and Kuljis, 2003); meso-level entity networks, including keyword networks (Radhakrishnan *et al.*, 2017); and micro-level entity networks, including dataset networks (Yu *et al.*, 2015), method networks (Li and Yan, 2018).

Knowledge networks facilitate observing the scientific world based on knowledge unit connections and have contributed to research into discipline development, knowledge flow, and knowledge discovery (Huang *et al.*, 2021). Using network centrality to describe the node distribution, researchers can understand whether the cooperation between authors (Kumar, 2019), countries (Kark *et al.*, 2009), or institutions (Han *et al.*, 2014) is close, and whether there is knowledge linkage and exchange between different disciplines or fields (Song *et al.*, 2023). Network attributes also bring new ideas to traditional frequency-based entity evaluation. Ding *et al*. (2009) ranked the importance of scholars in a co-citation network based on the PageRank algorithm. Fiala (2012) defined the co-authorship network of the most prominent computer scientists. The same method was also used to identify key journals (Zhang *et al.*, 2009) and topics (Takeda and Kajikawa, 2008) and optimize knowledge retrieval and recommendation.

Some scholars considered the dynamic changes of networks that determine the network's final structure and topological properties (Abbasi *et al*., 2012; Dorogovtsev & Mendes, 2002). Liu and Xia (2015) analyzed the development of cooperation between interdisciplinary fields. Liu and Guan (2015) further explored the influence of collaborative capacity, network status position, and cohesion on the growth and diversity of cooperative networks. Zheng *et al*. (2017) focused on the community in the network; they measured the academic influence of a community through the authors' influence. In addition to a single network, multiple networks were also constructed to discuss the relationship between their evolutionary trends. Singh *et al*. (2020) analyzed the evolution of a citation network and a co-author network to analyze the relationship between the distance in the co-author network and the citation pattern in the citation network.

Although network analysis has been widely used in research into knowledge entities, the research objects have generally been macro-level and meso-level knowledge entities, such as journals, papers, scholars, and keywords. A small number of micro-knowledge entity networks, such as software networks, only explain the influence of nodes and communities in static networks. As a new research object in knowledge entity analysis, algorithm entities have not yet benefited from network analysis. Therefore, we built a paper-level co-occurrence network of algorithms based on the full text of papers, and analyzed the academic influence based on the network features.

## 3. Methodology

We demonstrated the process of extracting algorithm entities from academic papers and leveraging their co-occurrences to create diverse networks. We then proceeded to conduct evaluations of the impact of these algorithmic nodes within these networks.

### 3.1 Data collection

The conference proceedings of the Annual Meeting of the Association for Computational Linguistics (ACL) conference were selected for algorithm identification. We chose these papers because scholars in the NLP field generally publish new research as conference articles, and algorithms mentioned in the ACL conference are more representative since ACL is one of the top NLP meetings. Additionally, the ACL conference is held annually and provides conference papers from its inception to the present day, allowing us to understand the historical development of NLP research and methods.

We used deep learning methods to identify algorithms, and the collected data could be classified into a raw corpus and a training corpus. Wang and Zhang (2020) manually annotated the algorithm entities mentioned in the full text of ACL conference papers from 1979 to 2015, and obtained sentences mentioning algorithms in the full text of papers. We preprocessed their algorithm sentences from each paper and collected 65,000 sentences in 4,604 papers as our training corpus.

ACL conference papers published from 2016 to 2020 were chosen as the raw corpus for the extraction of new algorithm entities, and these papers in PDF format were downloaded from the ACL Anthology website (https://aclanthology.org/venues/acl/). We invited five annotators to copy the PDF documents manually, then saved the content as TXT documents, and added the tag for each paper's title, authors, section, and paragraph. Finally, we obtained 2,793 papers in TXT format as the raw corpus.

### 3.2 Extraction of algorithm entities from NLP papers

As shown in Figure 1, identifying algorithm entities from the full-text content of articles is regarded as a named entity recognition (NER) task. We began by training various models and recognizing candidate entities from the raw corpus. The new algorithm entities were selected from candidate entities and combined with the existing entities in the training corpus to compile the algorithm dictionary. Finally, we reconfirmed the articles that mention each algorithm by dictionary matching, because the extraction model might miss an algorithm in a paper, but the algorithm could be found again by dictionary matching.

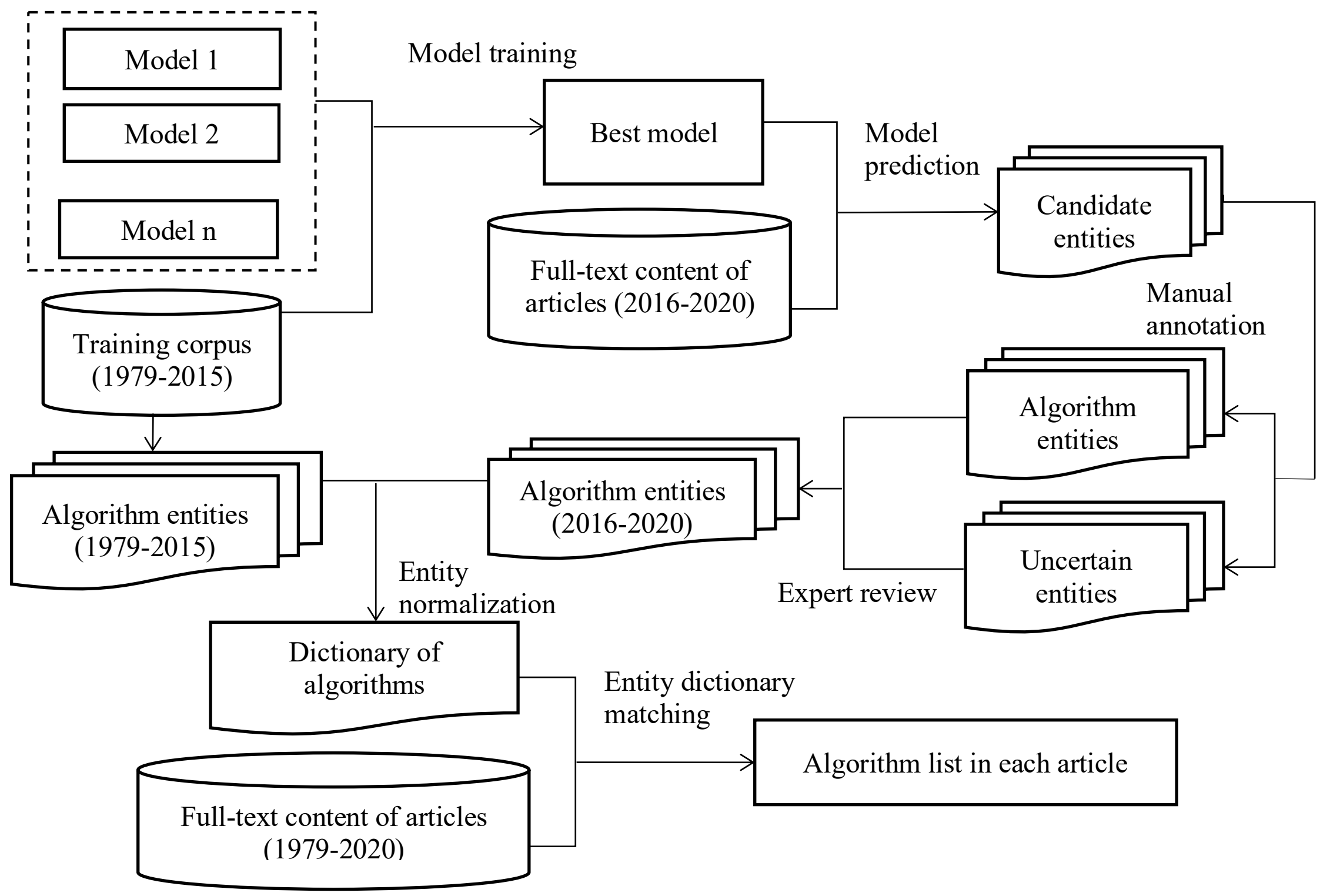

**Source(s): Figure by authors**

**Figure 1. Flow chart of the process of identifying algorithm entities in papers**

### 3.2.1 Extraction of candidate algorithm entities

The extraction of method entities from academic papers, such as algorithm entities, can be regarded as a named entity identification (NER) task (Hou *et al.*, 2022). Three popular deep learning models were chosen and compared for the preliminary extraction, namely BiLSTM+CRF (Panchendrarajan and Amaresan, 2018), BERT+CRF (Souza *et al.*, 2020), BERT+BiLSTM + CRF (Dai *et al.*, 2019). The training corpus consisted of 65,000 sentences, and each was annotated with the algorithm entities that appeared in that sentence. We divided the corpus into train-dev-test datasets by a 70%-15%-15% ratio and performed Hyper-parameter tuning for three extraction models. We used a BERT-base pre-trained language model, which consists of 12 layers, 768 hidden size, and 12 attention heads. Hyper-parameter tuning was conducted for each model. The setting of the main hyper-parameters can be seen in Table 1, and the other hyper-parameters followed by Devlin *et al*. (2019) and Dai *et al*. (2019).

**Table 1. Hyper-parameter setting**

**Source(s): Table by authors**

| Hyper-parameter | Range |
|---|---|
| Batch size | 8, 16, 32 |
| Learning rate (with linear scheduling) | 2e-5, 3e-5, 5e-5 |
| LSTM layers | 1, 2 |
| LSTM unit | 256, 512, 1024 |
| Dropout | 0.1 |
| Max epochs | 50 |

Table 2 presents the performance of each model where the BERT+BiLSTM+CRF got the highest $F_1$ score. Thus, we chose the BERT+BiLSTM+CRF model to predict candidate algorithm entities. From 472,470 sentences in 2,793 full-text publications, we obtained 300,839 sentences in 2,719 papers that contained at least one candidate algorithm entity.

**Table 2. Performance of the three extraction models**

**Source(s): Table by authors**

| Model | Precision | Recall | $F_1$ value |
|---|---|---|---|
| **BiLSTM+CRF** | 0.872 | 0.901 | 0.8862 |
| **BERT+CRF** | 0.953 | **0.962** | 0.9574 |
| **BERT+BiLSTM+CRF** | **0.970** | 0.959 | **0.9642** |

### 3.2.2 Identification of the algorithm entities

We manually labeled the candidate entities to identify the exact algorithm entities. First, we removed 16,108 candidate words that appeared in only one sentence. From the remaining 13,711 candidate words, we randomly selected 10% (1,371 words) to test the feasibility of manual labeling. A Ph.D. candidate and graduate students were invited to independently label these words and identify algorithm entities by the following rules:

(1) For a well-known algorithm, or if all of the authors of the ACL papers called the words an "algorithm," it was directly labeled as an "algorithm," such as support vector machines.

(2) For new words created by ACL authors, the labeler needed to read the definition in the original text and determine whether it was an algorithm.

(3) For a noun that the author did not define in the paper, for example, the Gibbs sampling method, the annotator searched for the noun in external knowledge bases—Wikipedia, Google Scholar, other academic papers, or monographs—and determined whether it was an algorithm.

(4) The remaining words were marked as "uncertain words" and required further judgment.

The interrater reliability (IRR) between the two labelers was 0.76 (Cohen's kappa coefficient), indicating that one labeler annotating all papers was sufficiently reliable. Therefore, the Ph.D. student annotated the remaining words, and obtained 1,576 algorithm entities and 622 uncertain entities.

For the uncertain entities, a professor familiar with NLP reviewed them and recognized 412 words as algorithm entities. In summary, we obtained 1,988 independent algorithm entities from 2,719 academic papers, which were then merged with algorithm entities in the training corpus. After deleting duplicate records, we collected 3,841 algorithm entities from all ACL papers and assigned them individual IDs.

### 3.2.3 Normalization of algorithm names

Authors are creative when mentioning algorithms in their papers. First, an algorithm could be mentioned in as many as a dozen forms. For example, "support vector machine" is also called "SVM," "SVMs," "support vector machines," "support-vector machine," or "support-vector machines." We grouped and normalized all names of an algorithm to prevent them from being multiple-counted by different names. Second, an entity word may also represent different algorithms. For example, "the BP algorithm" can represent either the "back-propagation" algorithm or the "belief propagation" algorithm. Third, an entity word may denote algorithms and some other common nouns. For example, "EM" can represent either the EM algorithm or the M entity ($e_1$, $e_1$, …, $e_m$).

For the first problem, we categorized all the algorithm words into two classes, full-name words, and acronyms, based on word length and manual review. For full-name words, we used the edit distance algorithm to assess the similarity between different names. If the similarity exceeded 95%, we considered them references to the same algorithm. We then manually reviewed these sets and chose the longest word as the standard name for the algorithm. A full-name dictionary containing 1,632 groups with the standard name and alias was compiled.

Abbreviations cause the other two problems. We extracted the original sentence mentioning the acronym and found the full-name words that appeared in the same sentence. We first determined whether these full-names and acronyms represented the same algorithm by two methods: firstly, the full-name word appeared in brackets after the acronym or vice versa, and secondly, the full-name words had the same combination of initial letters as the acronyms, and then we conducted a round of manual review. If the rule-based approach failed to identify full-name, matching was done by manually checking the original text. An acronym dictionary was constructed in which the standard name corresponding to each acronym were categorized into groups.

#### 3.2.4 Matching algorithm entities in each paper

We used a dictionary-based matching method to ensure the algorithm entity in each article by the full-name and acronym dictionary. When a full name appeared in a paper, we used their ID to replace the original name because there was no ambiguity. However, for abbreviations, we followed a different process:

(1) Using the abbreviation dictionary, we first identified the articles in which the abbreviation and the full name appeared simultaneously. By checking which full name appeared with the abbreviation, the abbreviation in the original sentence could be replaced with the ID of the full name.

(2) For the remaining articles containing abbreviations but not full names, if the acronyms were mentioned in more than two sentences, we manually assessed whether the abbreviations represented algorithms and their full name. Otherwise, we ignored the abbreviations with low frequency.

Finally, after de-duplication, we extracted the algorithm IDs in each article and obtained the algorithm entity list that appeared in each paper.

### 3.3 Analysis of algorithm influence based on co-occurrence network

#### (1) Algorithm co-occurrence network construction

The algorithm co-occurrence network was constructed based on co-occurring algorithm pairs in the same article. Each algorithm was considered as a node, and an edge was created between two nodes if they appeared in the same paper. We used the number of articles mentioning two algorithms to weigh the edges. The total number of ACL papers published each year was utilized to normalize the number of articles mentioning algorithms pair, and then the sum of normalized results each year was regarded as the edge weight.

Based on the above strategy, we constructed three kinds of networks. The first was the overall algorithm co-occurrence network, built based on the co-occurrence in all ACL papers. The second was the cumulative networks, and they were constructed by collecting the co-occurring algorithm pairs from 1979 to each year. The third one was the annual network. We extracted the co-mentioned pairs from papers published each year and built the annual networks.

#### (2) The influence of algorithms based on the co-occurrence network

The existing research has demonstrated that metrics based on network centrality are effective in analyzing influence (Ding *et al*., 2009). With the availability of full-text academic paper data, networks built on papers, keywords in papers, and knowledge entities have gradually emerged, and various centralities are also used to uncover important knowledge entities (Li and Yan, 2018), essential keywords (Lozano *et al.*, 2019), core terms (Grames *et al.*, 2019), and representative research topics (Yuan *et al.*, 2022). Therefore, after demonstrating that algorithm networks are consistent with the characteristics of complex networks, this study also used network centrality as the metric and analyzed the academic influence of algorithm entities from various perspectives. Regarding the selection of centrality metrics, we first considered some commonly used classic indicators, which have been employed in many studies for evaluating nodes such as papers, entities, and keywords in networks (Kim et al., 2014; Kumar and Srivastava, 2021;

Li and Yan, 2018; Ortega, 2021). At the same time, we also chose newly proposed indicators to maintain the completeness of our metric framework. The selected centralities assessed the influence from four aspects: popularity in the field, control within the group, central position within the group, and the balance of its influence. The specific interpretations of each indicator are as follows.

**Popularity:** We used Degree Centrality (DC) from the complex network to represent the popularity of individual algorithms within a group. As the simplest and most direct indicator of importance, the degree centrality of a node can be measured by the number of neighbors, which reflects the popularity and activity of the node (Freeman, 1979). The higher the degree centrality of an algorithm, the more relationships and stronger cohesion it has.

**Control:** The control of the individual algorithm in the group was represented by Betweenness centrality (BC), which was calculated by the portion of the number of shortest paths that pass through the given node (Brandes, 2001). Unlike degree centrality, BC measures not the direct connections that the node has in the population, but the ability to mediate among nodes that are not directly linked between them(Freeman, 1980). Therefore, the BC indicates "the potential of a point for control of communication" (Freeman, 1977). The higher the BC of an algorithm, the better its ability to connect different modules throughout the network, since members in the middle of a pathway have a more significant interpersonal influence on members at both ends of the path (Otte and Rousseau 2002).

**Central position:** Closeness centrality (CC) (Dangalchev, 2006) was used to reflect the centrality or central position of algorithm nodes within a group. CC finds the shortest path between each node first, then assigns a score to each node according to the sum of path length. The CC is a proxy for the independence and efficiency of communicating with other nodes; therefore, it helps identify strategically positioned nodes within the network (Abbasi *et al.*, 2012). For an algorithm, the higher its closeness, the closer it is to other algorithms, and the geometric center in the network. Therefore, the algorithm can most efficiently obtain and disseminate information through the network.

**Balance:** we used the Adjacency information entropy (AC) (Xu *et al.*, 2020) to measure the balance of the algorithm's impact on the network. The AC calculated the degree of node V (Kv) and the degree of all neighbor nodes of one node connected to V (Aj), and the entropy of proportion of Kv in Aj represented the influence of V in the network. The larger AC means the more balanced degree of node V in the social circle of all its neighbor nodes, and does not make up an extreme position in one neighbor's social circle.

## 4. Results

In this chapter, we offer comprehensive responses to the research question posed in the previous section. Our results begin by presenting the properties of the overall network, which is constructed based on the co-occurrence of all the algorithms. Subsequently, we provide the influence of algorithms based on different networks. Finally, we delve into the evolution of impact concerning individual algorithms within the network, supported by an in-depth analysis of the contributing factors to these trends.

### 4.1 Overview of algorithm co-occurrence networks

We answered the first research question in section 4.1. From the 7,397 papers, we extracted 1,632 algorithm entities, among which 6,488 papers (87.71%) mentioned at least one algorithm in the main text. We constructed the overall co-occurrence network of all algorithms. The average clustering coefficient, the relative size of the giant component, and the average shortest path of the network are 82.5%, 99.2%, and 2.225, respectively, indicating a small world of algorithms in the NLP domain. Next, we analyzed the degree distribution in the network, where degree represents the number of algorithms that are mentioned together with the target node. The degree distribution, P(K), is the probability that a randomly selected algorithm node has K edges. As illustrated in Figure 2, the degree distribution displays a power-law tail, known as the scale-free network. This observation signifies the existence of an importance gap or wealth gap between algorithms, and a small number of nodes are involved in most of the links in the network.

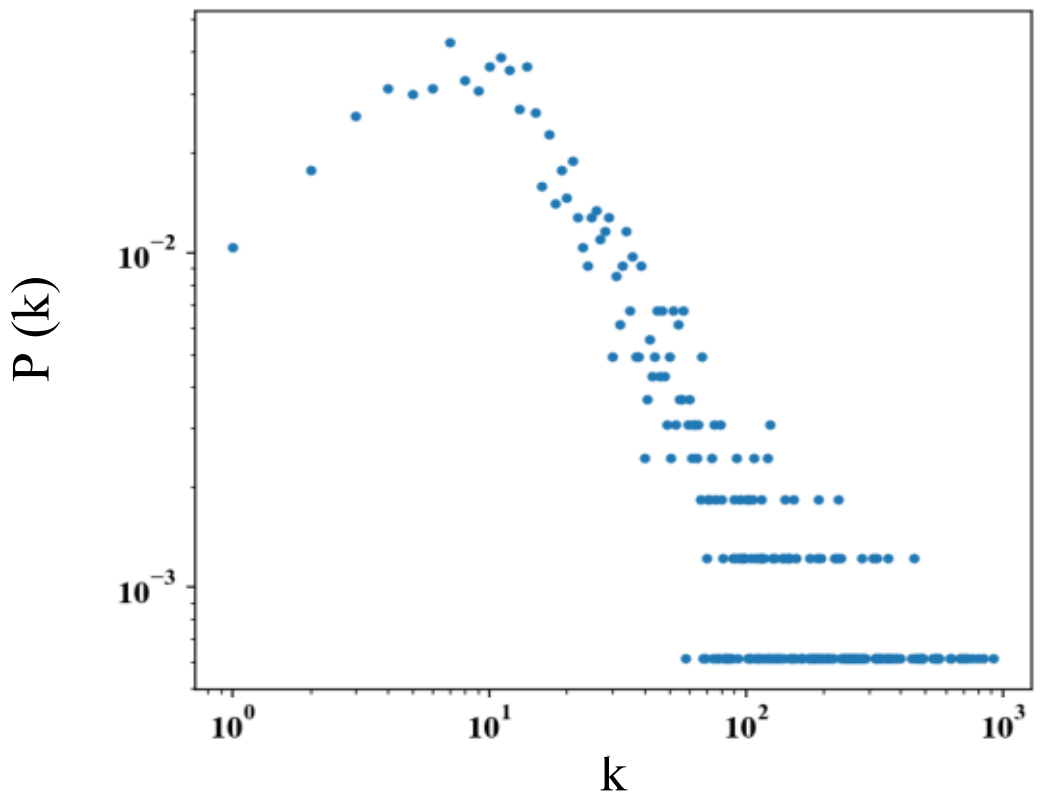

**Source(s): Figure by authors**

**Figure 2. Degree distribution of the co-occurrence network of algorithms**

Based on the findings above, we observe that algorithm co-occurrence networks share similarities with other complex networks, such as scholarly collaboration networks and actor networks (Albert and Barabási, 2000). They exhibit notably short distances between nodes, significant clustering coefficients, and a degree distribution that resembles a power-law. Consequently, we posit that the metrics traditionally employed to gauge node importance in complex networks can also be applied to assess the significance of algorithms within algorithm co-occurrence networks.

Furthermore, we analyzed how the current algorithm co-occurrence network was formed and what evolution it underwent. Specifically, we drew 41 cumulative co-occurrence networks based on the co-mentions of the algorithm in papers from 1979 through each year. Figure 3 shows the average degree of all nodes in each network, a metric that characterizes the network's interconnectedness. The rising average degree signifies the strengthening interconnections among algorithms, with the year 2001 marking a significant turning point in growth rates.

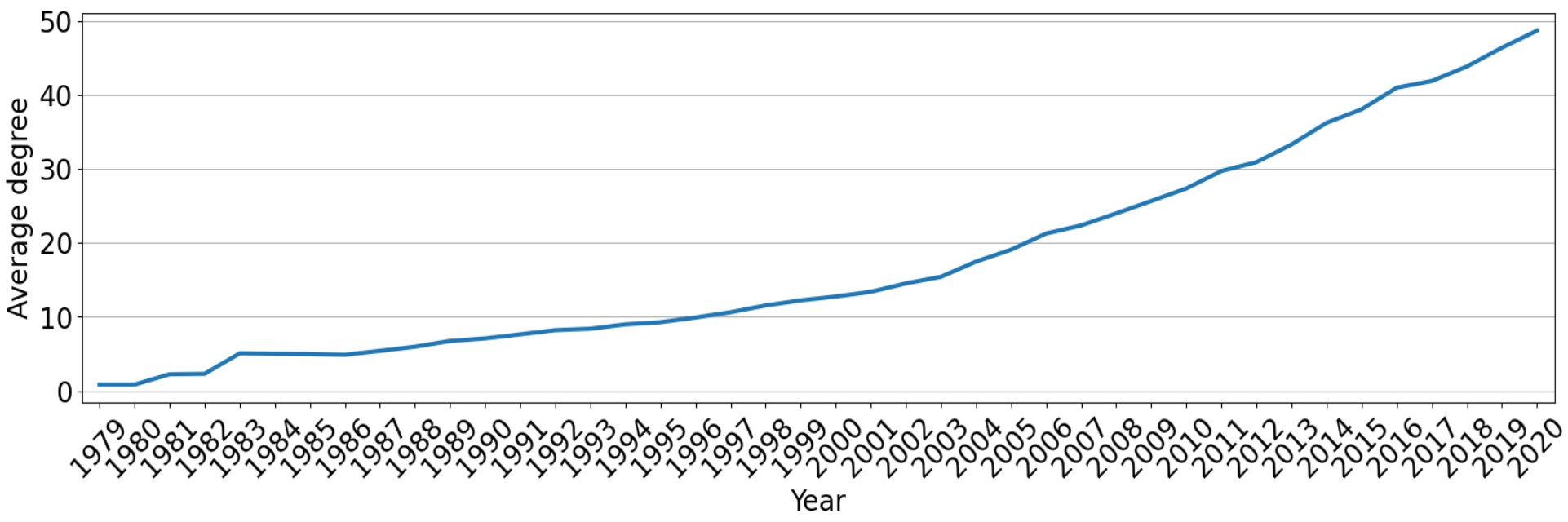

**Source(s): Figure by authors**

**Figure 3. The average degree of cumulative algorithm co-occurrence network**

The average length of the shortest path between each node pair is called the average separation. In Figure 4, the average separation experienced a substantial increase, peaking in 1994, followed by a steady decrease. This pattern suggests that the paths between algorithm pairs became longer as new algorithms continued to enter the network. After 1994, a more closely connected "small world" was gradually formed. Despite the expansion in network size and the number of algorithm nodes, the distance between algorithm pairs decreased, leading to improved communication efficiency.

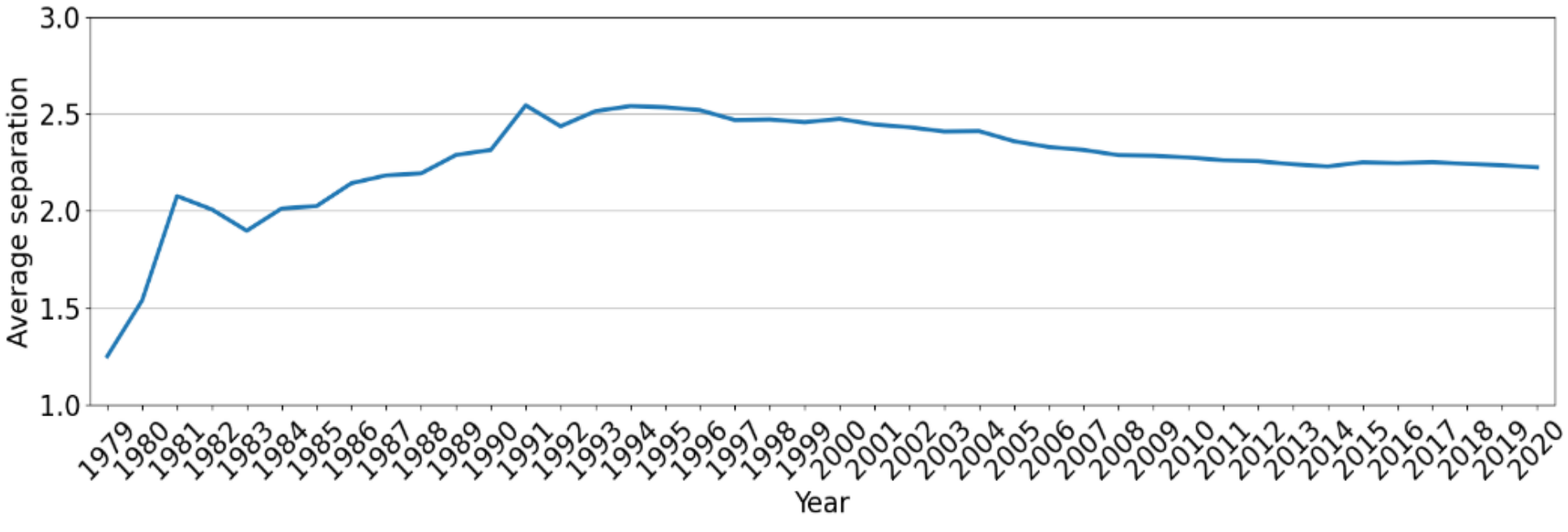

**Source(s): Figure by authors**

**Figure 4. Average separation distribution of the cumulative algorithm co-occurrence network by year**

The clustering coefficient reflects the degree of connection between all neighbor nodes of a focal node. Figure 5 illustrates the changes in the average clustering coefficient of the cumulative network, and data from 1979 and 1980 are excluded since the value is zero. From 1981 to 1998, the clustering coefficient showed pronounced fluctuations, changing approximately every five years. After reaching a trough in 1998, the clustering coefficient rose steadily, experienced a flat period between 2008 and 2012, and rose again until the last year. These fluctuations in the early network highlight its unstable structure, but the consistent trend aligns with the evolution of the average separation, demonstrating the network's increased connectivity in recent years.

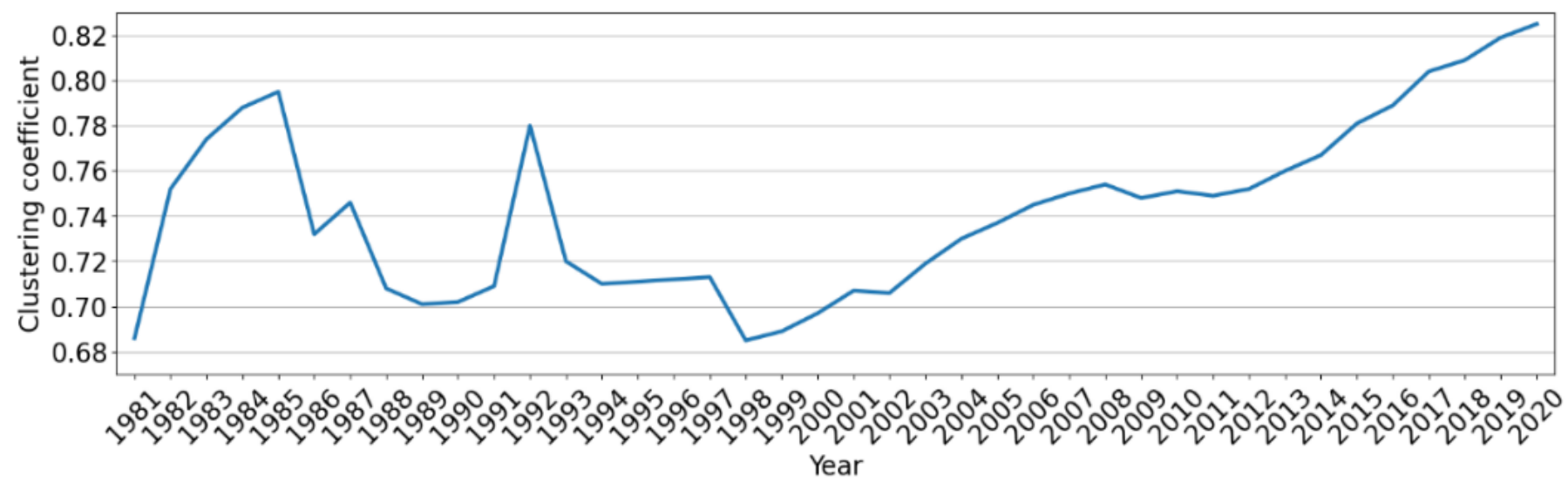

**Source(s): Figure by authors**

**Figure 5. The clustering coefficient of the cumulative algorithm co-occurrence network by year**

Figure 6 presents the relative size of the giant component in each cumulative network. The relative size increased markedly in the first five years, and after a brief decline from 1983 to 1986, it rose sharply again until 1993. From 1993-2006, the relative size gradually exceeded 95%. It can be speculated that the topology of the algorithm network stabilized after 2006, and almost all algorithms appeared in the giant component.

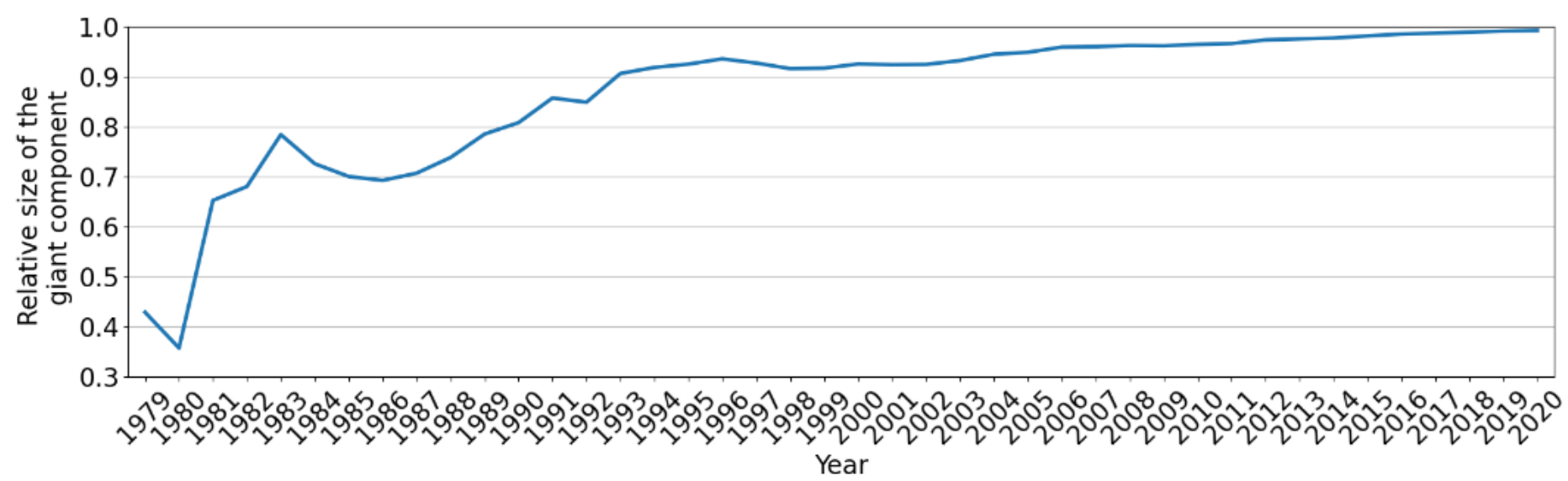

**Source(s): Figure by authors**

**Figure 6. The relative size of the giant component in the cumulative algorithm co-occurrence network by year**

The above results show that the close correlation between algorithms in the field of NLP did not happen overnight. In the first 20 years, the structure of co-occurring networks formed between algorithms was not tight. After almost 20 years of turbulence, the algorithm network entered a period of stability and formed a small world.

## 4.2 The influence of algorithm based on the co-occurrence network

In section 4.2, we answered the second research question. Specifically, we explored the academic influence of each algorithm in the overall algorithm network and each year's algorithm network separately to see if there are differences in the roles of algorithms in the network.

### (1) Algorithms with high influence in overall co-occurrence network

We conducted an in-depth analysis of the overall co-occurrence network to derive multiple centrality measures encompassing popularity, control, central position, and balance, for each algorithms. For a more intuitive comparison of the different algorithms, we refer to the work of Wang and Zhang (2020) and categorize the algorithms into different types, including Classification algorithm, Clustering algorithm, Dimension reduction algorithm,

Gramar, Integrated learning algorithm, Link analysis algorithms, Metric algorithm, Neural networks, Optimization algorithm, Probabilistic graphical models, Regression algorithm, Search algorithm, Unique algorithms in NLP domain. We ranked the algorithms in the same category according to their academic impact scores in different dimensions. Table 3 presents the top five algorithms in each algorithm type, but we only show the results of algorithms in four categories because of the limited space.

Overall, different indicators do not yield the same results in ranking the influence of algorithms, regardless of the algorithm category. Therefore, it is valuable to measure the popularity, control, central position, and balance of individual algorithms in the community by different network centralities, respectively. Comparing the algorithms that rank high on different dimensions of influence, it can be found that the ranking results of "control" are different, where algorithms that do not appear in other lists appeared in the top list of BC, for example, word2vec, random walk, and viterbi algorithm. Furthermore, the algorithms with the highest scores for all indicators in different categories point to the same algorithm, and we try to analyze why these best-performing algorithms achieve the highest scores.

Among different neural network models, Long Short-Term Memory (LSTM) emerged as the algorithm with the highest influence. As a special recurrent neural network (RNN) that introduced the "gating mechanism", LSTM is widely regarded as one of the most successful RNN models and has achieved state-of-the-art results on various sequence modeling tasks. LSTM is often co-mentioned with similar algorithms related to the "gating mechanism" (Sahu and Anand, 2016), frequently used with the probabilistic graphical model, namely, CRF, to ensure accurate sequence ordering in output, and compared or combined with grammatical algorithms (Liu et al., 2018; Stanojević and Steedman, 2020). Consequently, LSTM has more neighboring algorithms and is closely related to other algorithms. At the same time, LSTM has the most balanced influence over its neighbors, i.e., it doesn't have an extraordinarily large or small impact over just one of them.

**Table 3. The top-5 algorithms with high influence based on each network indicators**

**Source(s): Table by authors**

| Type | Rank | Popularity | Control | Central position | Balance |
|---|---|---|---|---|---|
| Classification algorithm | 1 | Support vector machine | Support vector machine | Support vector machine | Support vector machine |
| | 2 | Maximum entropy | Maximum entropy | Maximum entropy | Maximum entropy |
| | 3 | Gibbs sampling | Decision tree | Decision tree | Gibbs sampling |
| | 4 | Decision tree | Gibbs sampling | Naive bayes | Decision tree |
| | 5 | Naive bayes | Naive bayes | Gibbs sampling | Naive bayes |
| Neural networks | 1 | Long short-term memory | Long short-term memory | Long short-term memory | Long short-term memory |
| | 2 | Recurrent neural network | Recurrent neural network | Recurrent neural network | Recurrent neural network |
| | 3 | Bi-LSTM | Glove | Bi-LSTM | Glove |
| | 4 | Glove | Bi-LSTM | Glove | Bi-LSTM |
| | 5 | Convolutional neural network | Word2vec | Convolutional neural network | Convolutional neural network |
| Optimization algorithm | 1 | Adaptive moment | Adaptive moment | Adaptive moment | Adaptive moment |
| | 2 | Stochastic gradient descent | Dynamic programming | Stochastic gradient descent | Stochastic gradient descent |
| | 3 | Dynamic programming | Stochastic gradient descent | Dynamic programming | Dynamic programming |
| | 4 | Expectation maximization | Expectation maximization | Maximum likelihood estimation | Expectation maximization |
| | 5 | Maximum likelihood estimation | Viterbi algorithm | Expectation maximization | Maximum likelihood estimation |
| Probabilistic graphical models | 1 | Conditional random field | Conditional random field | Conditional random field | Conditional random field |
| | 2 | Hidden markov model | Hidden markov model | Hidden markov model | Hidden markov model |
| | 3 | Latent dirichlet allocation | Latent dirichlet allocation | Latent dirichlet allocation | Probabilistic latent semantic analysis |
| | 4 | Probabilistic latent semantic analysis | Random walk | Probabilistic latent semantic analysis | Latent dirichlet allocation |
| | 5 | Baysian network | Baysian network | Baysian network | Baysian network |

The algorithm that takes on the most influential in classification algorithms is the support vector machine (SVM). As described in the "Top 10 algorithms in data mining" (Wu et al., 2008), the SVM algorithm has a solid theoretical foundation. It is one of the most stable and accurate algorithms among all well-known algorithms. Compared with other contemporaries, SVM exhibits exceptional generalization capabilities, proficiently tackling classification and regression challenges involving high-dimensional features, and it was regarded as the best algorithm among traditional machine learning algorithms. It can be speculated that in academic papers, SVM not only needs to be compared with other statistical machine learning classification algorithms, e.g., naive bayes, decision tree, but also needs to be used as a baseline model to validate the superiority of emerging deep learning algorithms, e.g., BERT, on classification tasks. Therefore, SVM can have a multitude of co-mentioned neighboring nodes, and a higher betweens and a core position as a transitional algorithm from statistical machine learning algorithms to deep learning algorithms.

The algorithms that rank high in the other categories also possess the advantages that LSTM and SVM have within their respective types. The above discussion indicates that when we consider the interconnections among algorithms and examine their popularity, control, central position, and balance within the entire community, algorithms that score highly across multiple indicators share some common characteristics. They not only have a history but also have not been eliminated by history, allowing them time to accumulate popularity. Their emergence in the field of NLP often coincides with the junction of different eras, enabling them to compare with old and new algorithms and thus act as a bridge controlling the community's connections. These algorithms also possess superior performance and high versatility across various tasks, which places them at the core of the community and endows them with a more balanced academic influence.

**(2) Algorithm with high influence in annual co-occurrence network**

Based on the annual co-occurrence network, we obtained the influence of the algorithm each year by different centrality. Table 4 shows the most important algorithms in the network each year. Regardless of the era, algorithms that score the highest on various indicators are generally the same, indicating that when an algorithm acquires the most neighboring nodes in the network, it also becomes the central hub of the entire network, serving as a bridge for communication among different algorithms. During the first 14 years, grammatical algorithms were prominent in the field. Three of the most vital grammar class algorithms, ATN, CFG, TAG, have the same function, but ATN and TAG are improved based on CFG and have more powerful performance, so the three always alternate in the grammar era as the most critical nodes in the network. From 1994, statistical machine learning algorithms like the hidden markov model (HMM) gained the highest compromise centrality, yet context-free grammar (CFG) remained in the top position.

**Table 4. Algorithms with the highest influence each year based on network indicators**

**Source(s): Table by authors**

| Year | Popularity | Control | Central position | Balance |
|---|---|---|---|---|
| 1979 | ATN | ATN | ATN | discrimination network |
| 1980 | ATN | ATN | ATN | greedy regression |
| 1981 | ATN | ATN | ATN | left corner |
| 1982 | ATN | ATN | ATN | ATN |
| 1983 | CFG | CFG | CFG | CFG |
| 1984 | LFG | ATN | LFG | DCG |
| 1985 | CFG | ATN | CFG | CFG |
| 1986 | CFG | GPSG | GPSG | GPSG |
| 1987 | CFG | CFG | CFG | CFG |
| 1988 | TAG | CFG | CFG | TAG |
| 1989 | TAG | DCG | DP | TAG |
| 1990 | CFG | CFG | CFG | CFG |
| 1991 | CFG | CFG | CFG | CFG |
| 1992 | TAG | TAG | TAG | CFG |
| 1993 | CFG | CFG | CFG | CFG |
| 1994 | CFG | HMM | DCG | left corner |
| 1995 | CFG | MLE | MLE | CFG |
| 1996 | CFG | DP | CFG | CFG |
| 1997 | MLE | MLE | decision tree | MLE |
| 1998 | DP | DP | HMM | HMM |
| 1999 | DP | DP | DP | DP |
| 2000 | HMM | MLE | HMM | MLE |
| 2001 | MaxEnt | MaxEnt | MaxEnt | MaxEnt |
| 2002 | MaxEnt | decision tree | MaxEnt | MaxEnt |
| 2003 | MaxEnt | CFG | MaxEnt | MaxEnt |
| 2004 | SVM | SVM | HMM | SVM |
| 2005 | MaxEnt | SVM | MaxEnt | MaxEnt |
| 2006 | MaxEnt | SVM | MaxEnt | MaxEnt |
| 2007 | MaxEnt | SVM | MaxEnt | MaxEnt |
| 2008 | SVM | SVM | SVM | SVM |
| 2009 | SVM | SVM | SVM | SVM |
| 2010 | SVM | SVM | SVM | SVM |
| 2011 | SVM | SVM | SVM | SVM |
| 2012 | SVM | SVM | SVM | SVM |
| 2013 | SVM | SVM | SVM | SVM |
| 2014 | SVM | SVM | SVM | SVM |
| 2015 | SVM | SVM | SVM | SVM |
| 2016 | RNN | word2vec | RNN | word2vec |
| 2017 | LSTM | LSTM | LSTM | LSTM |
| 2018 | LSTM | LSTM | LSTM | LSTM |
| 2019 | LSTM | LSTM | LSTM | LSTM |
| 2020 | BERT | BERT | BERT | BERT |

**Note**: Abbreviation in the table. ATN (augmented transition network); BERT (bidirectional encoder representations from transformer); CFG (context free grammar); DP (dynamic programming); GPSG (generalized phrase structure grammar); HMM (hidden markov model); LFG (lexical functional grammar); LSTM (long short term memory); MaxEnt(maximum entropy); MLE (maximum likelihood estimation); SVM (support vector machine); TAG (tree adjoining grammar ); RNN(recurrent neural network).

From 1997 to 2015, statistical machine learning algorithms continued to dominate, with SVM, HMM, and MaxEnt being the most representative. Indeed, an iterative relationship exists among these three algorithms. HMM is employed for large-scale real-text processing but cannot consider contextual features during the feature selection process. The emergence of MaxEnt effectively compensates for this deficiency, as MaxEnt can utilize arbitrarily complex features. Simultaneously, MaxEnt can tag individual words but does not take advantage of the relationships between tokens, but HMM excels in establishing markov correlations between tokens. Consequently, these two algorithms function independently and can be synergistically combined to form a maximum entropy markov model (MEMM). In contrast to other algorithms, SVM lies in its ability to effectively address classification and regression problems involving high-dimensional features, a challenge that proves formidable for many other algorithms. 2016 was the year when the algorithm network transitioned from traditional machine learning to deep learning. With RNN gaining the biggest influence in the network, the most representative RNN, namely LSTM, has become the algorithm with the most neighbors, the highest influence on neighbors, and the most core position in the network. However, the monopoly of LSTM was broken in 2020, and BERT, proposed less than two years ago, soon became a star in the algorithm network, indicating that the pre-training model is gradually replacing the traditional neural network model.

### 4.3 The evolution of algorithms influence based on the co-occurrence network

We answered the third research question in section 4.3. Based on the above results, we identified three eras in developing the NLP field. We selected the most representative algorithms that obtained the highest influence within their category of each period and analyzed the evolution of different types of academic influence.

#### (1) The influence evolution of context-free grammar

CFG was selected as a representative of grammar algorithms. As shown in Figures 7(a) and 7(b), CFG had a high influence from 1979 to 1998, consistently ranking in the top ten in terms of the number of neighbors or its core position in the network, and scores of each influence exceeding 0.5 were at the high level of its entire life cycle. As early as the 1960s, Chomsky proposed CFG, which was powerful enough to express the grammatical patterns of most programming languages (Kadlec, 2008). After more than ten years of development, CFG gained a high reputation in the algorithm community due to the small number of algorithms in the early years. However, CFG's simplicity made it challenging to describe languages with complex features. Consequently, scholars put forward various constraints to optimize CFG, including augmented transition network (ATN), tree adjoining grammar, and functional unification grammar (FUG). Generally, between 1979 and 1998, CFG can be regarded as the originator of grammatical algorithms, spawning many new algorithms, and being compared with other grammar, which brought it many algorithm neighbors and a central position.

After 1998, the influence of CFG experienced slight fluctuations and then entered a recession period in the ranking. The period can be divided into two stages: 2003-2012 and 2013-2020, with the latter stage experiencing a stronger downward trend than the former. Especially between 2003 and 2007, the NLP field shifted from rule-based rationalism to statistics-based empiricism. Grammatical algorithms such as CFG were no longer suitable for handling massive and complex natural languages. The rise of machine learning and deep learning algorithms led to a sharp decrease in the influence of CFG, and the elimination of its related algorithms. Consequently, its influence continued to decline, with the number and importance of nodes connected to it constantly decreasing.

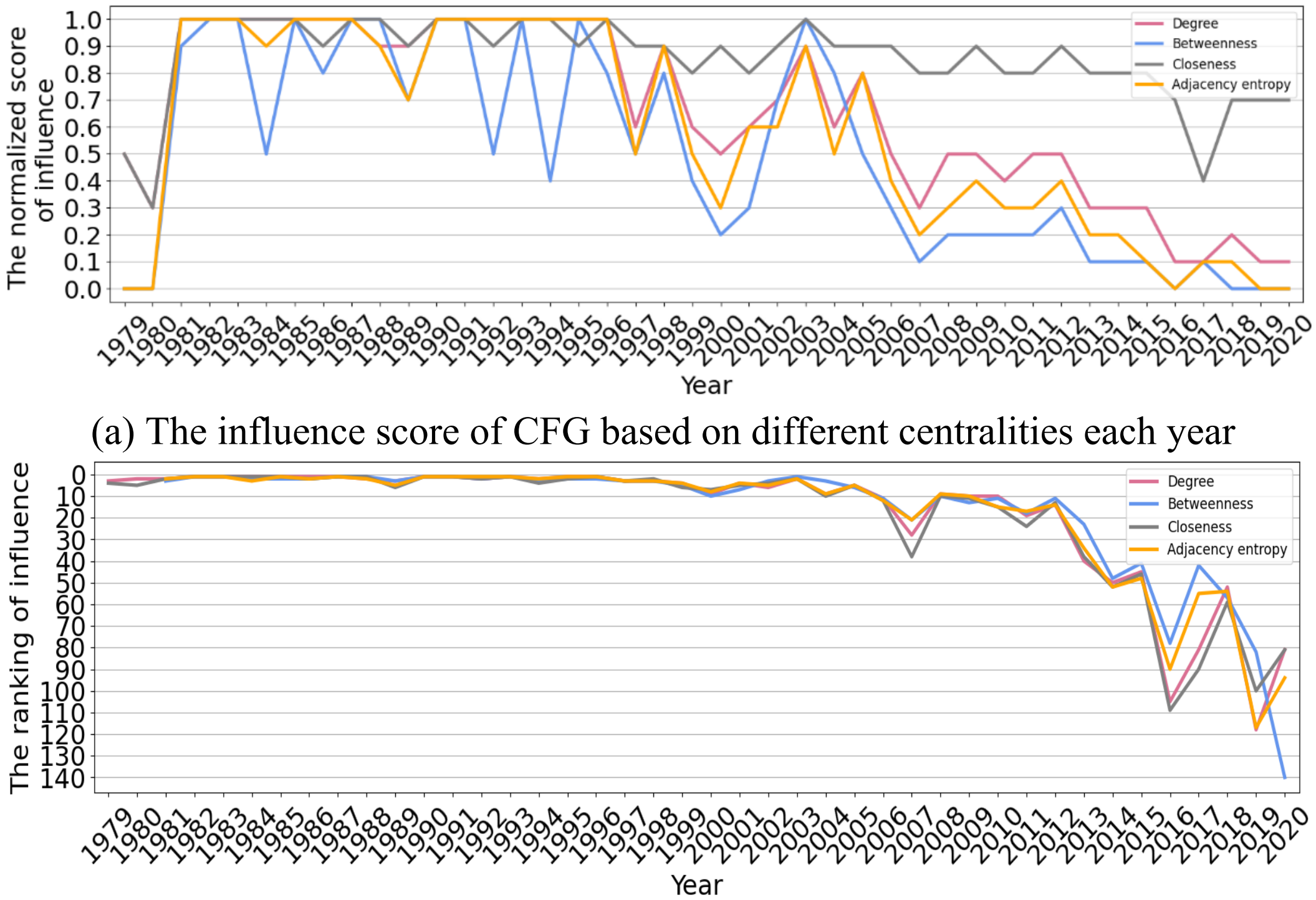

(a) The influence score of CFG based on different centralities each year

(b) The influence ranking of CFG based on different centralities each year

**Source(s): Figure by authors**

**Figure 7. The influence evolution of context-free grammar**

**(2) The influence evolution of support vector machine**

From the statistical machine learning algorithms, we choose the support vector machine (SVM), and its evolution based on influence score and ranking is shown in Figure 8 (a) and (b), respectively. SVM first appeared in the ACL paper in 2000. Unlike CFG, the influence of SVM in the network did not reach its peak as soon as it appeared. Since only five years had passed between the SVM being used and its appearance in ACL, it was not yet connected with other algorithms to achieve a higher group status. Since 2002, the influence ranking of SVM began to increase significantly. In 2004, SVM ranked first and got the highest impact scores in its academic career. Our analysis of the tasks handled by SVM over the years revealed a steady expansion in the range of tasks. In 2000, SVM was only used for discourse block recognition. Then, it solved the text clustering task in 2001, and by 2004, it began to process semantic disambiguation, parsing, sentiment classification, automatic summarization, relationship recognition, and many other tasks. The wider the application range of SVM, the more opportunities it must cooperate and compare with other algorithms, and the faster the influence would grow.

From 2005 to 2014, SVM entered a boom period, during which its centrality rankings fluctuated slightly but remained consistently in the top ten, and its centrality scores never declined. In this decade, many derivative algorithms and tools were proposed, such as the well-known LibSVM, and the basic SVM would appear together with more new algorithms. After 2015, the scores and rankings began to drop sharply, because deep learning algorithms began to be widely used. SVM was not the core problem-solving method but appeared more as a baseline method. While SVM is the most representative statistical machine learning algorithm, deep learning algorithms have replaced its core position. Deep learning algorithms may have also negatively affected other statistical machine learning algorithms. As a result, the number and influence of algorithms adjacent to SVM also decrease, causing a drop in DC and AC scores.

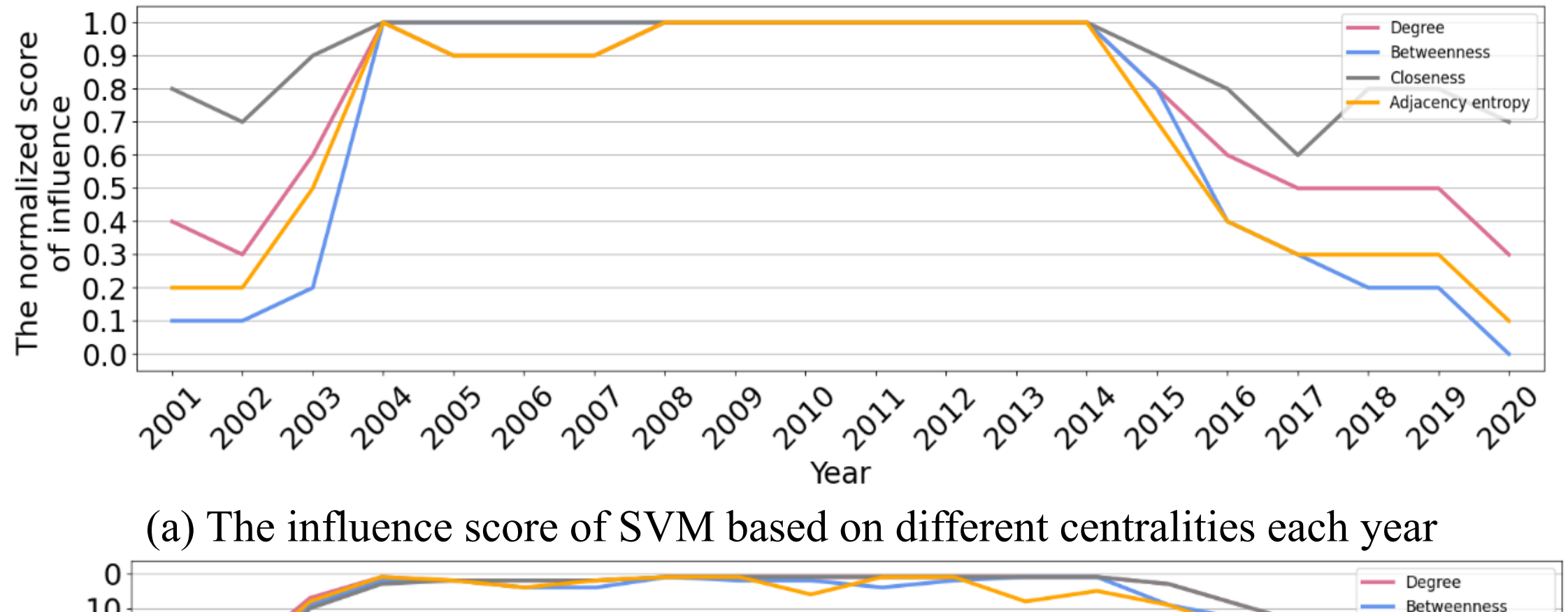

(a) The influence score of SVM based on different centralities each year

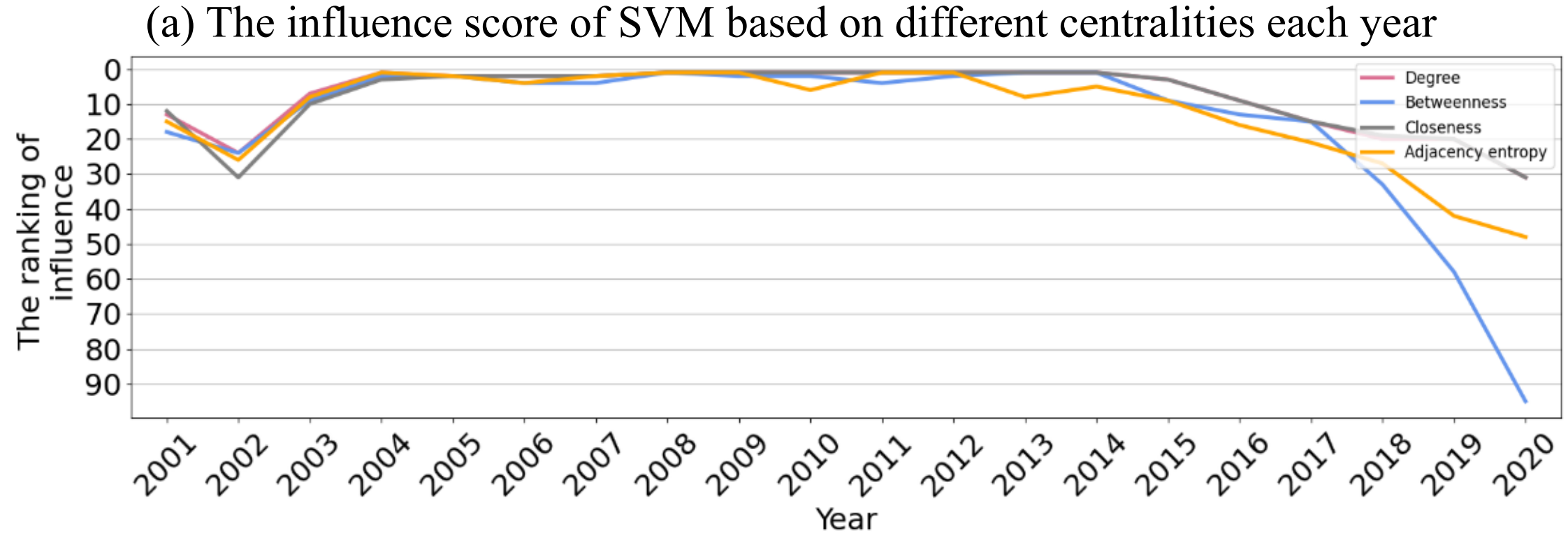

(b) The influence ranking of SVM based on different centralities each year

**Source(s): Figure by authors**

**Figure 8. The influence evolution of support vector machine**

**(3) The influence evolution of long short-term memory**

We selected long short-term memory (LSTM) as the representative deep learning algorithm. LSTM was initially proposed in 1997 (Hochreiter and Schmidhuber,1997). In the first ten years of the 21st century, scholars proposed some epoch-making models, such as LSTM with a forget gate structure, but it didn't garner substantial attention and widespread adoption in NLP until 2010 (Van Houdt *et al.*, 2020). As depicted in Figure 9(a), LSTM was first introduced at the ACL conference in 2013, when the centrality rankings were quite low. However, the network influence has grown extremely fast, ranking in the top 10 of all algorithms in each score within just three years. In 2015, Google first tried to combine LSTM with CTC (connectionist temporal classification) to process speech text. In the same year, two famous variants of LSTM, highway networks and gated recurrent unit (GRU), were proposed, and bidirectional LSTM was also combined with CRF for sequence annotation. Technology giants such as Google, Facebook, and Apple used LSTM to process sequence labeling tasks and applied it in various software such as translation software, voice input, and subtitle generation. Mentioning these new methods would simultaneously introduce the source algorithm, LSTM, which could explain why its network influence ranking peaked after 2015.

The influence of LSTM began to decline in 2019. The decline in the ranking may be due to the increase of new algorithms, but the decrease in the score shows that other algorithms have shaken the position of LSTM. At the end of 2018, the pre-training model BERT was proposed and set off a new round of deep learning model revolution in the NLP field. Various algorithms derived from BERT are also emerging in an endless stream, and LSTM and related algorithms are inevitably affected. Therefore, each influence score and ranking of LSTM have shown a clear downward trend.

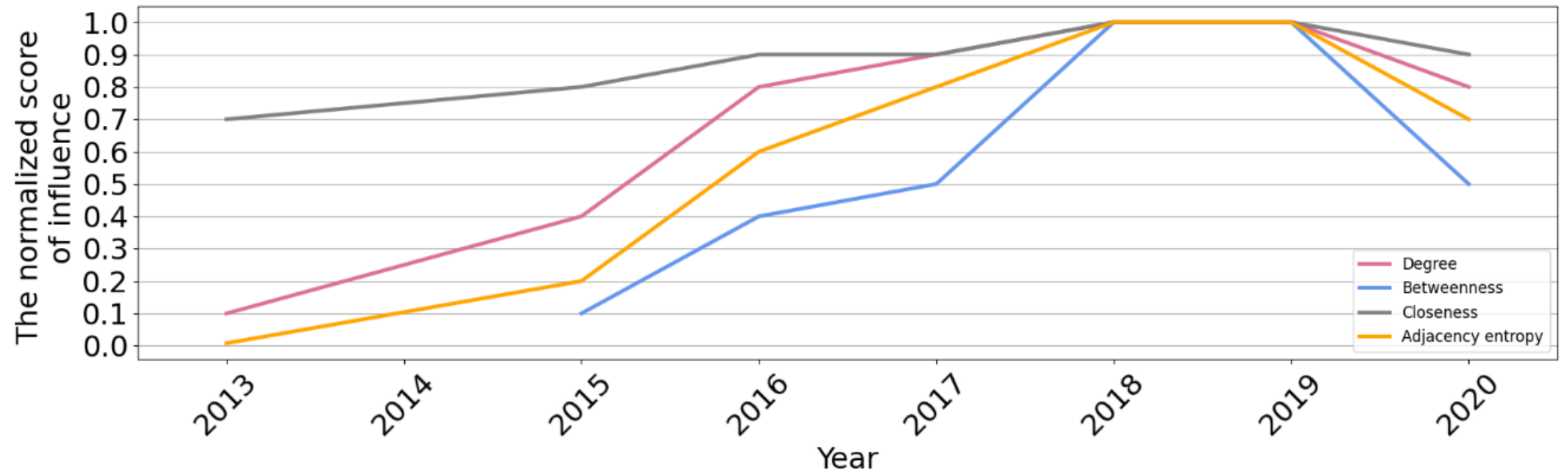

(a) The influence score of LSTM based on different centralities each year

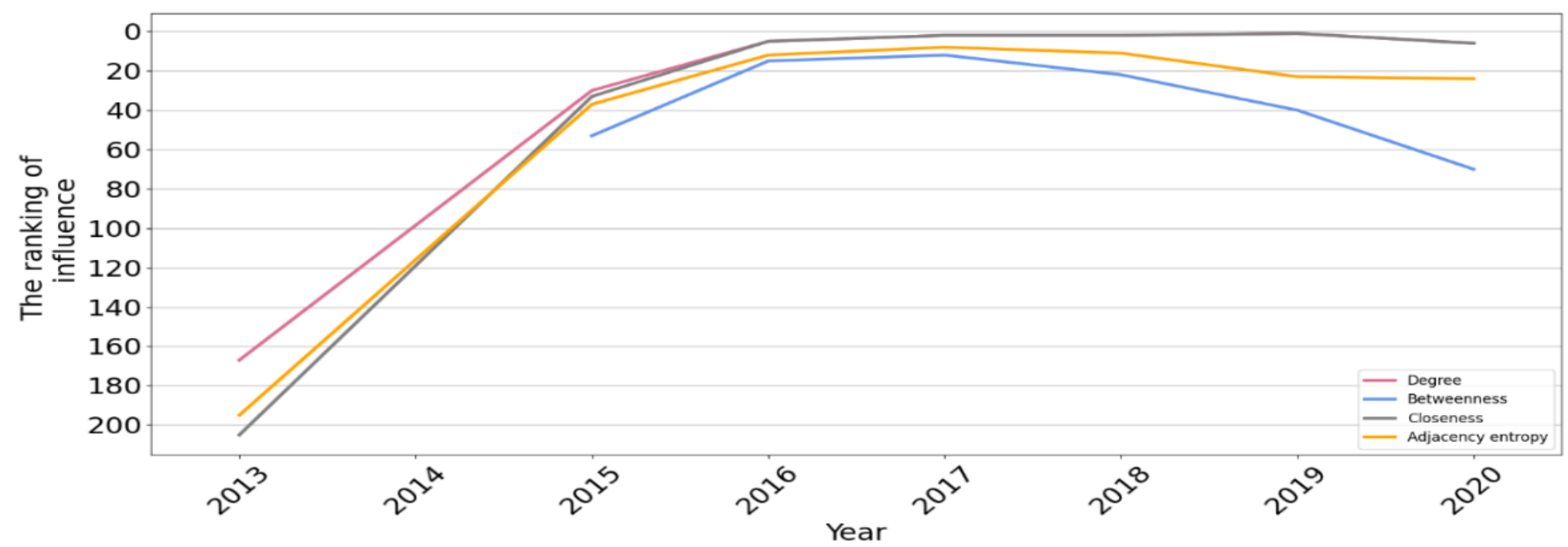

(b) The influence ranking of LSTM based on different centralities each year

**Source(s): Figure by authors**

**Figure 9. The influence evolution of long short-term memory**

## 5. Discussion

**(1) Influence ranking of algorithms based on traditional frequency indicator**

Following the existing method (Wang & Zhang, 2020), we calculated the academic influence of algorithms based on the number of papers mentioning them. Table 5 presents the top 10 algorithms ranked according to their frequency-based influence, where the most influential one is the newly proposed algorithm (compared to the data collection time), BERT. However, when we compare the findings in Table 5 with our network-based results in the previous section, a compelling trend emerges: algorithms with a longer history get a significant ascent in rankings when we account for their properties in the co-occurrence network.

Newly proposed algorithms often attract considerable attention due to their exceptional initial performance, leading to a notable academic impact that can be captured by the frequency-based method. Nevertheless, the frequency fails to comprehensively understand their vital roles and contributions within the broader research community, which are precisely the classical algorithms that stand out. The interconnected network of algorithms, as illuminated in this study, furnishes us with a unique opportunity to uncover hidden dimensions of value among classical algorithms. By utilizing this approach, we discern their influence and contributions to community development and categorize their specific roles in the evolutionary process. For instance, the support vector machine (SVM) emerges as a prominent "bridge builder," facilitating connections between other algorithms. This observation underscores the multifaceted roles algorithms assume in advancing the field. In light of these insights, the network-based evaluation approach introduces a novel dimension that augments our ability to discern algorithmic roles and influences that remain elusive when solely relying on traditional frequency-based metrics.

**Table 5. The top algorithms with high influence by frequency**

**Source(s): Table by authors**

| Rank | Algorithm |
|---|---|
| 1 | BERT |
| 2 | long short-term memory |
| 3 | transformer |
| 4 | bi-directional long short-term memory |
| 5 | support vector machine |
| 6 | word2vec |
| 7 | convolutional neural network |
| 8 | context free grammar |
| 9 | gated recurrent unit |
| 10 | maximum entropy |

### (2) Why do algorithms co-occur in papers in the NLP domain?

Our research illuminates a pronounced phenomenon within the realm of Natural Language Processing (NLP): a burgeoning level of interaction among algorithms within the network, a trend that exhibits growth over time. To unravel this intricate web of algorithm interactions, we systematically identified and scrutinized the edges with the highest weights within the annual network, thereby dissecting the algorithm nodes linked by these pivotal connections. Based on the top three edges of each year in Table 6, we can infer that the algorithms were mentioned together for three reasons. First, algorithms are conjoined due to a subordinate relationship, implying that it becomes imperative to introduce their parent algorithms when discussing sub-algorithms. For instance, "C4.5-Decision Tree," where C4.5 represents a specialized variant within the broader category of decision trees. Second, there is a competitive relationship between algorithms. When confronted with specific NLP tasks, researchers often assess and compare the performance of multiple algorithms of the same genre. For example, the decision tree and the support vector machine were used simultaneously to compare which performs better. Third, a cooperative relationship among algorithms comes into play, where employing two or more algorithms enhances outcomes compared to individual deployment. For example, "HMM-MaxEnt", combining HMM and MaxEnt can get MEMM and produce better results.

**Table 6. Top three algorithm pairs in the algorithm co-occurrence network by year**

**Source(s): Table by authors**

| Year \ Rank | 1 | 2 | 3 |
|---|---|---|---|
| 1979 | ATN-Discrimination Network | CFG-Local Maxima | |
| 1980 | CFG-PSG | ATN-Semantic Grammar | ATN-Author Topic |
| 1981 | Left corner parsing-Semantic grammar | ATN- Left corner parsing | CYK- Left corner parsing |
| 1982 | ATN-CFG | ATN-GPSG | ATN-CFPSG |
| 1983 | CFG-GPSG | GPSG-LFG | CFG-LFG |
| 1984 | CFG-LFG | DCG-LFG | CFG-DCG |
| 1985 | CFG-DCG | CFG- Left corner parsing | DCG- Left corner parsing |
| 1986 | CFG-TAG | CFG-GPSG | LFG-TAG |
| 1987 | CFG-DCG | DCG-FUG | CFG-FUG |
| 1988 | CFG-TAG | GPSG-TAG | CYK-CFG |
| 1989 | DCG- Left corner parsing | HDPSG-UCG | CFPSG- Left corner parsing |
| 1990 | CFG-TAG | CYK-CFG | CYK-TAG |
| 1991 | CFG-TAG | FUG-HDPSG | Left Corner-TAG |
| 1992 | CYK-CFG | CFG-TAG | CFG- Left corner parsing |
| 1993 | HMM-Mutual Information | CFG-HDPSG | CFG-Mutual Information |
| 1994 | CFG- Left corner parsing | DCG- Left corner parsing | DP-HMM |
| 1995 | Decision Tree-DP | Mutual Information-Yarowsky | HDPSG-TAG |
| 1996 | CFG-TAG | CYK-CFG | DTG-TAG |
| 1997 | EM-Maximum Likelihood Estimation | CFG-TAG | DCG-HDPSG |
| 1998 | CFG-TAG | C4.5-Decision Tree | HDPSG-TAG |
| 1999 | C4.5-Decision Tree | CFG-PCFG | Mutual Information-Nearest Neighbor |
| 2000 | HMM-MaxEnt | Decision Tree-MaxEnt | Decision Tree-HMM |
| 2001 | HMM-MaxEnt | DP-MaxEnt | CYK-DP |
| 2002 | HMM-MaxEnt | Generalized Iterative Scaling-MaxEnt | Decision Tree-MaxEnt |
| 2003 | HMM-MaxEnt | IBM Model-Viterbi Alignment | Beam Search-MaxEnt |
| 2004 | CFG-PCFG | DP-PCFG | EM-PCFG |
| 2005 | MaxEnt-SVM | HMM-MaxEnt | CRF-HMM |
| 2006 | Decision Tree-SVM | MaxEnt-SVM | Decision Tree-MaxEnt |
| 2007 | HMM-MaxEnt | CRF-HMM | MaxEnt-SVM |
| 2008 | Decision Tree-SVM | Naive Bayes-SVM | MaxEnt-SVM |
| 2009 | CRF-MaxEnt | MaxEnt-SVM | DP-HMM |
| 2010 | EM-HMM | CRF-HMM | MaxEnt-SVM |
| 2011 | MaxEnt-SVM | EM-HMM | Naïve bayes-SVM |
| 2012 | Gibbs Sampling-LDA | EM-HMM | MaxEnt-SVM |
| 2013 | Gibbs Sampling-LDA | CRF-SVM | CRF- Maxent |
| 2014 | Gibbs Sampling-LDA | Back Propagation algorithm-Neural Network | EM-HMM |
| 2015 | Neural Network-SGD | Neural Network-Softmax | Softmax-SGD |
| 2016 | Neural Network-RNN | LSTM-Neural Network | LSTM-RNN |
| 2017 | LSTM-Neural Network | Neural Network-RNN | LSTM-RNN |
| 2018 | LSTM-Neural Network | LSTM-RNN | BiLSTM-LSTM |
| 2019 | BiLSTM-LSTM | LSTM-Neural Network | LSTM-RNN |
| 2020 | Bert-Neural Network | Bert-Softmax | LSTM-Neural Network |

**Note:** Abbreviation in the table: ATN (augmented transition network); CFG (context free grammar); DP (dynamic programming); DTG (description tree grammar ); PCFG (probabilistic context free grammar); FUG (functional unification grammar); PSG (phrase structure grammar); UCG (unification categorial grammar); LFG (lexical functional grammar); TAG (tree adjoining grammar ); CYK (cocke younger kasami); HDPSG (head-driven phrase structure grammar); CFPSG(context free phrase structure grammar); GPSG(generalized phrase structure grammar); DCG (definite clause grammar); SVM(support vector machine); HMM (hidden markov model); EM (expectation maximization); LDA (latent dirichlet allocation); CRF (conditional random field); Maxent (maximum entropy); SGD (stochastic gradient descent); LSTM (long short term memory); BiLSTM (bi-directional long short term memory ); RNN(recurrent neural network).

**(3) The evolution rules of network influence**

We compared the influence ranking and evolution of different algorithms based on the network. Despite the inherent variability in the influence of individual algorithms, we identified several overarching trends that provide valuable insights into the dynamics of algorithmic influence.

Firstly, we observe that algorithms with a more extended utilization history, preceding their introduction to the ACL stage, tend to exhibit higher levels of network influence. Specifically, CFG took approximately 20 years, SVM six years, and LSTM three years from widespread use to being mentioned in ACL conferences. Moreover, the network centrality rankings and scores they received after their first appearance in ACL conferences gradually decreased, with CFG being the highest and LSTM the lowest. We hypothesize that if the algorithm had been used sufficiently before used in the ACL, it could have built up associations with other algorithms and accumulated enough reputation to gain a high network influence.

Second, the social relationships of algorithms at different career stages have changed. If an algorithm's social connections remain stable, its influence score does not change, but the exit of old algorithms and the addition of new algorithms may cause fluctuations in ranking. However, in our results, the network influence score of an algorithm also shows clear periods of rise, boom, and bust over time, implying that both the quantity and quality of connections with other algorithms evolve as the algorithm matures. When an algorithm enters a prosperous period, more and more neighbors connect with it, and its core position in the algorithm network becomes stable. Conversely, as it enters a period of decline, the number and significance of connections with other algorithms diminish, resulting in the gradual disintegration of its social relationships.

Third, the evolution of influence based on each centrality shows different trends. Notably, Betweenness Centrality always has the lowest score and the highest decline rate, while Closeness Centrality has the lowest decline rate. We postulate that as an algorithm's influence wanes within the group, it typically follows a progression: firstly, it no longer occupies the core position as a bridge; next, it gradually loses the proportion of influence among neighbor algorithms, and then the number of algorithms connected with it decreases continuously, while the algorithm's distance from all other algorithms changes minimally.

## 6. Conclusions

In this work, we explore the influence of algorithms based on the algorithm co-occurrence network in the NLP domain. First, we automatically extract algorithms in academic papers, construct an algorithm network according to their co-occurrence, and explore the evolving characteristics of the algorithm network. We find that knowledge networks represented by algorithms have the same characteristics as other complex networks. A shift from dispersion to co-prosperity during the network evolution indicates a growing symbiotic relationship between algorithms and NLP research. Then, we evaluate the influence of algorithms by the characteristics of various co-occurrence networks. The results indicate regardless of the category to which the algorithms belong, influential algorithms within the algorithm community are often classical algorithms with numerous derivative counterparts, possessing sufficiently good performance to drive the development of algorithms in the same category. We conduct a temporal analysis of the influence trends for three representative algorithms and elucidate the underlying factors governing their evolution. We observe that shifts in social relations coincide with the development of algorithms, with algorithms expanding their social circles during boom periods and contracting during recessions. During this process, algorithms typically lose their core network position and then lose associated neighbor nodes.

Although our study revealed valuable insights, there are still some limitations. First, we only constructed the network based on algorithms mentioned in NLP papers, and the evaluation results might not hold in a different field, as each domain has its unique ways of mentioning and utilizing the same algorithms. However, the comprehensive research approach provided in this paper is still transferable and can be applied to data in other fields. Secondly, we only consider co-occurrence at the article level, which could not help us obtain a more fine-grained co-occurrence relationship between algorithms. Finally, the network only reflects co-occurrence relationships, and some semantic relationships, such as cooperation or competition, could not be directly reflected.

In the future, we plan to explore co-occurrence relationships at a more granular level, such as the sentence or section level, and study the semantic relationships between pairs of co-occurring algorithms. To this end, we can add semantic edges to the network to understand how the algorithms were mentioned together. We also intend to incorporate author information, topics, or tasks of the papers that mention the algorithms to construct a multi-layer network, analyze the relationships between algorithms, tasks, and scholars in the field, and explore algorithm-based partner recommendation, task-based algorithm recommendation, and other valuable applications.

## Acknowledgments

This study has received support from the National Natural Science Foundation of China (Grant No. 72074113, 72374103) and the Ministry of Education of the Republic of Korea and the National Research Foundation of Korea (NRF-2020S1A5B1104865).